# Sound Abstraction of Probabilistic Actions in The Constraint Mass Assignment Framework

**AnHai Doan Peter Haddawy**
Decision Systems and Artificial Intelligence Laboratory
Department of EE & CS
University of Wisconsin-Milwaukee
Milwaukee, WI 53201
{*anhai, haddawy*}*@cs.uwm.edu*

## Abstract

This paper provides a formal and practical framework for sound abstraction of probabilistic actions. We start by precisely defining the concept of sound abstraction within the context of finite-horizon planning (where each plan is a finite sequence of actions). Next we show that such abstraction cannot be performed within the traditional probabilistic action representation, which models a world with a single probability distribution over the state space. We then present the constraint mass assignment representation, which models the world with a set of probability distributions and is a generalization of mass assignment representations. Within this framework, we present sound abstraction procedures for three types of action abstraction. We end the paper with discussions and related work on sound and approximate abstraction. We give pointers to papers in which we discuss other sound abstraction-related issues, including applications, estimating loss due to abstraction, and automatically generating abstraction hierarchies.

## 1 INTRODUCTION

Recently there have been a number of proposed techniques to reduce the running time complexity of probabilistic and decision-theoretic planning [2, 11, 8, 1, 7]. Most of these techniques are abstraction techniques or some variants of abstraction techniques. In Markov Decision Process planning, for example, Dean *et. al.*[2] plan with an incrementally increasing envelope of states, while abstracting all remaining states – those with low probability of being entered – into a single state. Boutilier *et. al.* [1] reduce the number of states by collapsing states differing only in the values of "irrelevant" domain attributes. In finite-horizon planning, the BURIDAN planner [11] employs abstraction techniques in one of its plan-evaluation procedures; it bundles action branches deemed to be similar in terms of the final outcomes. The DRIPS planner [7] uses a number of abstraction techniques to abstract plans and actions.

Most proposed abstraction techniques deal with one of two categories of abstraction: *approximate abstraction* and *sound abstraction*. Approximate abstraction gives approximate solutions that sometimes can be shown to be within an estimated distance from the optimal, exact solutions. Examples of approximate abstraction approaches are [2, 1]. These approaches suggest that approximate abstraction is used to reduce the size of the input planning problem by abstracting away unimportant detail. In contrast, sound abstraction tends to be employed only in some phases of the planning algorithm and aims at reducing the time necessary to find the optimal plans. The distinguishing characteristic of sound abstraction is that it guarantees the planner never jumps to false conclusions because inferences made at abstract levels must be consistent with those made at lower levels. Examples of sound abstraction approaches are BURIDAN [11] and DRIPS [7].

There has been little work attempting to formalize sound abstraction techniques for probabilistic planning despite the fact that such techniques are commonly used by several planners (e.g., BURIDAN, DRIPS), and that they are promising for Bayesian network inference, decision tree evaluation, as well as qualitative planning approaches. Developing a formal framework for sound abstraction involves several major tasks, including the following: First, a general representation framework must be developed, which defines the concepts of state, world, action, plan, projection, expected utility, etc. Within this framework, the concept of sound abstraction can then be precisely defined. Second, a concrete representation framework must be developed. This means giving well-defined syntax and semantics for worlds, actions, and plans; as well as developing procedures for projecting plans, and procedures for computing the expected utility of a plan. Third, within the concrete representation framework, sound abstraction types must be identified. Procedures for creating an abstract entity from member entities – such as creating the description of an abstract actions from the member actions – must be developed. And fourth, procedures for estimating information loss due to abstraction must be developed.

In a previous paper [8], we attempted to address the first three issues. We gave justifications for sound action abstraction. We then identified several useful types of action abstraction and presented procedures for creating the abstractions. But because we were working within the traditional probabilistic action representation (henceforth the SPD representation), which models a world with a single probability distribution, the abstraction techniques presented in that paper cannot be applied to abstract actions. This is a severe limitation since one expects to use such techniques to create abstraction hierarchies. So we need to find a new representation framework that accomodates sound abstraction, and to develop sound abstraction techniques within this framework.

This paper offers such a formal framework for sound abstraction of actions. First, we define sound abstraction in the context of finite-horizon probabilistic planning, where each plan is a finite sequence of actions. Second, we present the constraint mass assignment representation, which models a world with a set of probability distributions and is a generalization of mass assignment representations. Third, we present sound abstraction procedures for three types of action abstraction. Finally, we end the paper with a discussion of related work on sound and approximate abstraction. Due to space limitation, the reader is referred to [5] for a more detailed discussion of the topics of this paper, as well as discussions of applications, estimating loss due to abstraction, and automatically constructing abstraction hierarchies.

## 2 SOUND PROBABILISTIC ACTION ABSTRACTION

In this section we present a general representation framework for finite-horizon decision-theoretic planning. We discuss the relationship among states, worlds, actions, and plans. We also discuss the need for projection procedures and define sound projection procedures. Within this framework we define sound abstraction.

A *state* completely specifies the planning domain at a particular moment. We denote the set of all possible states with $\Omega$.

A *world* represents the uncertainty about the state of the planning domain at a particular moment. We interpret each world as representing a set of probability distributions over $\Omega$. Typically, worlds are represented using some representation mechanism that can not represent all possible sets of probability distributions.

A *concrete action* maps a set of probability distributions into a set of probability distributions. This mapping is represented with the function *exec(action, set of probability distributions)*. The equation $\wp_{post} = exec(\mathsf{a}, \wp_{pre})$ specifies that the action a maps the set of probability distributions $\wp_{pre}$ into the set of probability distributions $\wp_{post}$.

*Concrete plans* are finite sequences of concrete actions. Like actions, plans also map sets of probability distributions into sets of probability distributions. We represent this mapping with the function *exec(plan, set of probability distributions)*. For a plan $p = \mathsf{a}_1\mathsf{a}_2\cdots\mathsf{a}_n$, where the $\mathsf{a}_i$ are concrete actions, we have
$exec(p, \wp_{pre}) = exec(\mathsf{a}_n, exec(\mathsf{a}_{n-1}, \cdots, exec(\mathsf{a}_1, \wp_{pre})\cdots))$.

In executing a plan, computing the post-execution set of probability distributions exactly is typically impossible in all but very small domains. A common approach to solve this problem is to develop a projection procedure that yields a post-projection set of probability distributions that is an approximation of the post-execution set. Denote the mapping induced by this projection procedure with the function *project(action, world)*[1], the equation $w_{post} = project(\mathsf{a}, w_{pre})$ specifies that when the action a is projected in the world $w_{pre}$ (or rather on the set of probability distributions represented by the world $w_{pre}$) we obtain as the post-projection result the (set of probability distributions represented by) world $w_{post}$. Projecting a plan $p = \mathsf{a}_1\mathsf{a}_2\cdots\mathsf{a}_n$ in a world $w_{pre}$ yields a world $w_{post}$ that is computed as

$$\begin{aligned} w_{post} &= project(p, w_{pre}) \\ &= project(\mathsf{a}_n, \cdots, project(\mathsf{a}_1, w_{pre})\cdots). \end{aligned}$$

We say the projection procedure associated with the funtion *project* is *sound* if and only if for any concrete plan $p$ and world $w$ we have $project(p, w) \supseteq exec(p, w)$. An example of a sound projection procedure is the FORWARD procedure presented in [11].

An *abstract action* is built from concrete or lower-level abstract actions using a number of abstraction operators; each *concrete instantiation* of an abstract action is a sequence of concrete actions. An *abstract plan* is a finite sequence of abstract and concrete actions. If the abstract plan $p^*$ is $\mathsf{a}^*_1\mathsf{a}^*_2\cdots\mathsf{a}^*_n$, where the $\mathsf{a}^*_i$ are abstract or concrete actions, then a *concrete instantitation* of the plan $p^*$ is $s_1 s_2 \cdots s_n$, where each $s_i$ is a concrete instantiation of $\mathsf{a}^*_i$.

We say an abstraction framework that satisfies the above requirement is *sound* (and so are abstraction operators in that framework) if and only if for any world $w$, plan $p$ and its concrete instantiation, $p_i$, we have $project(p, w) \supseteq exec(p_i, w)$.

[1]The function *project* is defined for worlds because we are concerned only with the cases where plans are executed on sets of probability distributions that can be represented with worlds.

## 3 THE CONSTRAINT MASS ASSIGNMENT REPRESENTATION

In the previous section we presented a general representation framework for probabilistic planning in which worlds are represented with sets of probability distributions over the state space. In this section we introduce a concrete representation framework that will serve as our substrate for sound abstraction.

**Motivation**

We have to find a mechanism to represent worlds, that is, sets of probability distributions. A natural starting point is the SPD framework which is the underlying representation for the majority of existing probabilistic and decision-theoretic planning systems [11, 2] and is introduced in [11]. In this framework a world is modeled with a single probability distribution over the state space. Projecting a plan thus results in a probability distribution over the state space.

Unfortunately, the SPD framework does not accommodate sound abstraction as defined in Section 2. To see this, consider two actions $a_1$, $a_2$, and a world $w$ such that $exec(a_1, w) \neq exec(a_2, w)$. If we are to abstract these two actions into an abstract action $a^*$ then by the definition of sound abstraction (Section 2) we have $project(a^*, w) \supseteq exec(a_1, w)$ and $project(a^*, w) \supseteq exec(a_2, w)$. Because $exec(a_1, w) \neq exec(a_2, w)$ it follows that $project(a^*, w)$ represents at least two different PDs. The definition of the function *project* (Section 2) implies that $project(a^*, w)$ can be represented with a world. But in the SPD framework a world cannot represent two different PDs!

The next natural candidate mechanism for representing a set of probability distributions is mass assignments [12], which constitute a subclass of lower probabilities. Mass assignments are very appealing because they have the intuitive interpretation of assigning probability masses to sets of states and can be very compactly represented in practice. Unfortunately, we have been unable to develop effective projection rules and abstraction procedures for the mass assignment framework. To the best of our knowledge, most existing work on developing this framework presents projection rules that iterate through all subsets of the state space and so cannot be applied to domains with large state space. (For a discussion on this topic see [4, 5].) We have found that a generalized representation of the mass assignment framework, which we call the *constraint mass assignment* framework, is well-suited for practical projection rules and sound abstraction.

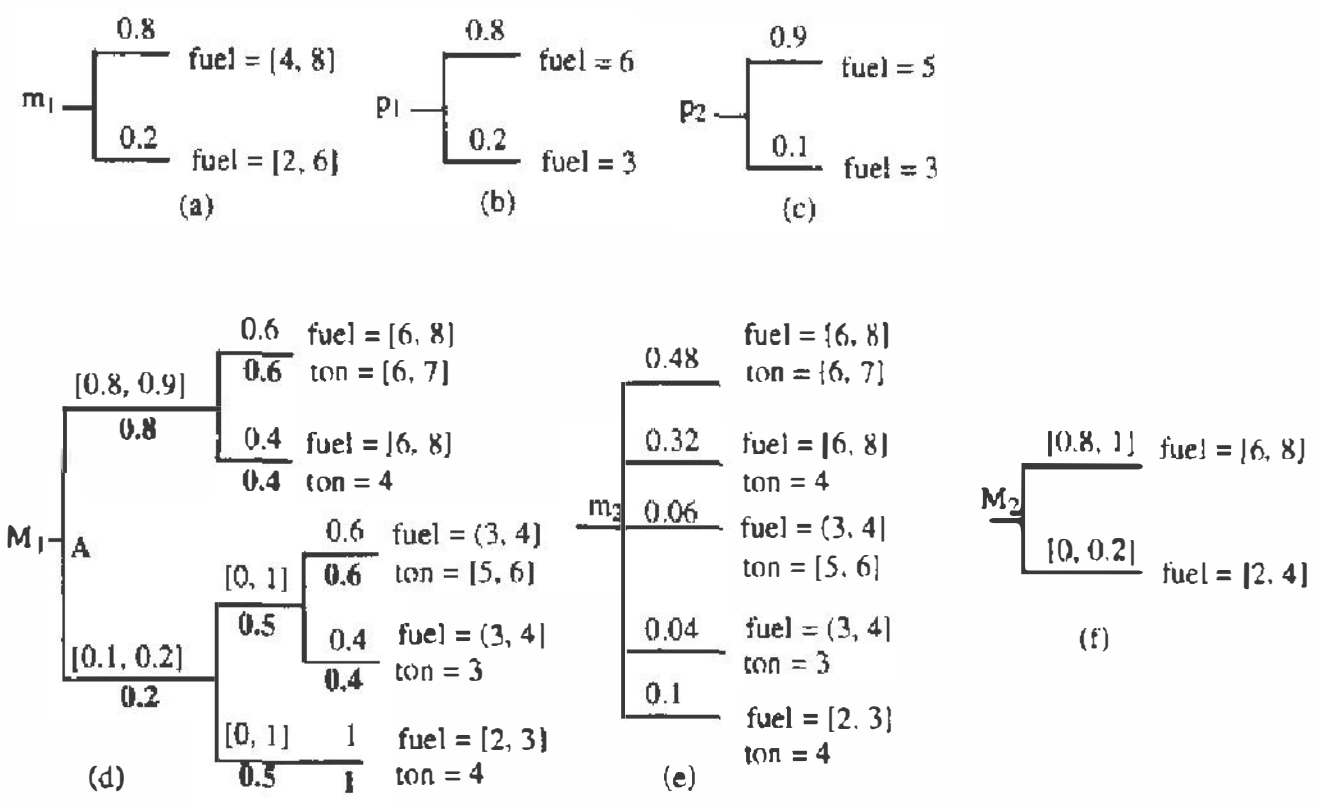

Figure 1: Examples of MAs, CMAs, and IMAs.

Since an understanding of mass assignments facilitates an understanding of constraint mass assignments, we start in the next subsection with a brief introduction to mass assignments

**Mass Assignments**

Let $\Omega$ denote the state space; a mass assignment (MA) $m: 2^\Omega \rightarrow [0, 1]$ assigns to each subset of $\Omega$ a portion of the probability mass [12]. We have $\sum_{B \subseteq \Omega} m(B) = 1$ and $m(\emptyset) = 0$. The set $B \subseteq \Omega$ is called a *focal element* of $m$ if $m(B) > 0$; and we will say the pair $\langle B, m(B) \rangle$ forms a *branch* of $m$. A probability distribution $P$ is said to be *consistent* with a mass assignment $m$ iff $P(B) = \sum_{b \in B} P(b) \geq \sum_{C \subseteq B} m(C)$ for all $B \subseteq \Omega$. The set of probability distributions consistent with $m$ will be denoted as $\wp(m)$.

A mass assignment $m$ can be interpreted as representing the uncertainty concerning the true probability distribution over $\Omega$ with a set of probability distributions, namely $\wp(m)$. Figure 1.a shows a mass assignment, the first (second) focal element of which consists of all the worlds in which *fuel* is between 4 and 8 (2 and 6). Note that these two focal elements are non-disjoint sets of states. Examples of the probability distributions consistent with this mass assignment are shown in Figures 1.b and 1.c.

**Constraint Mass Assignments**

Simply put, a constraint mass assignment (CMA) encodes a set of MAs using a tree representation. A CMA is specified with a tree each leaf of which is a subset[2] of $\Omega$ and each branch of which is assigned a real interval that is in [0,1] and is called *the branch's probability interval.* Figure 1.d shows the tree of a CMA $M_1$ that has depth three (for the moment ignore the numbers in bold fonts). The tree branches stemming from a node are called *the sibling branches* of each other.

[2] The leaf sets can be non-disjoint sets of states.

An MA $m$ that belongs to the set of MAs encoded by a CMA M (denoted as $m \in M$) can be generated as follows: (1) assign to each branch $br$ of the tree T representing M a number such that this number belongs to $br$'s probability interval and such that the sum of this number and all other numbers assigned to the sibling branches of $br$ is one; (2) for each path from T's root until a leaf $B_i$ obtain the product $P_i$ of the assigned numbers along the path; (3) form the mass assignment $m$ that has as its branches the pairs $< B_i, P_i >$, where the $B_i$ are the leaves of the tree T.

To illustrate the above generation process, consider again the CMA $M_1$ shown in Figure 1.d. We assign to each branch the number in bold font shown next to that branch. It is easy to check that the numbers on sibling branches sum to one. The MA $m_2$ obtained from assigning these numbers is shown in Figure 1.e.

The set of probability distributions represented by a CMA M, denoted by $\wp(M)$, is defined to be $\bigcup_{m \in M} \wp(m)$, where $\wp(m)$ is the set of probability distributions represented by the MA $m$. For the sake of simplicity, we will write $M$ to denote $\wp(M)$ when doing so does not introduce ambiguity. For example, instead of writing $\wp(M_{post}) \supseteq exec(\mathsf{a}, \wp(M_{pre}))$, we will write $M_{post} \supseteq exec(\mathsf{a}, M_{pre})$.

It is not difficult to see that a CMA – specified by giving its tree structure, branch probability intervals, and state sets representing the leaves – precisely defines a unique set of MAs; and an MA precisely defines a unique set of probability distributions. So CMAs are well-defined representations of sets of probability distributions.

CMAs with trees of depth one form a special subclass of CMAs that we term *the class of interval mass assignments (IMAs)*. Figure 1.f shows for example an IMA $M_2$.

In a related paper [4], we presented a framework for using IMAs to represent worlds and discussed in detail the advantages of the IMA framework over the MA framework. We pointed out the two key advantages: first, the IMA framework is more expressive than the MA framework, and second, the former can acommodate projection and abstraction techniques that the latter cannot. The CMA framework inherits all these advantages. In addition, the CMA framework is more expressive and can accommodate better projection techniques than the IMA framework [5]. Because of all the above theoretical and practical advantages, we model worlds with CMAs.

### Concrete Actions

A concrete action can be depicted with a tree structure as shown in Figure 2.a, where the $c_i$ are mutually exclusive and jointly exhaustive conditions on $\Omega$, the $I_{ij}$ are probability intervals in [0,1], and the $E_{ij}$ are the effects. Each effect $E_{ij}$ is a function mapping a state into a sets of states $E_{ij} : \Omega \rightarrow 2^{\Omega}$. Abusing notation, for a set $B \subseteq \Omega$ we will use $E_{ij}(B)$ to denote the set $\bigcup_{b \in B} E_{ij}(b)$. Figure 2.b shows an example action description in which the amount of goods delivered varies, depending on $fuel$ being greater or less than 3. An assignment such as $ton = ton + [2,3]$ states that $ton$ after executing the action will be increased by an unknown amount between 2 and 3.

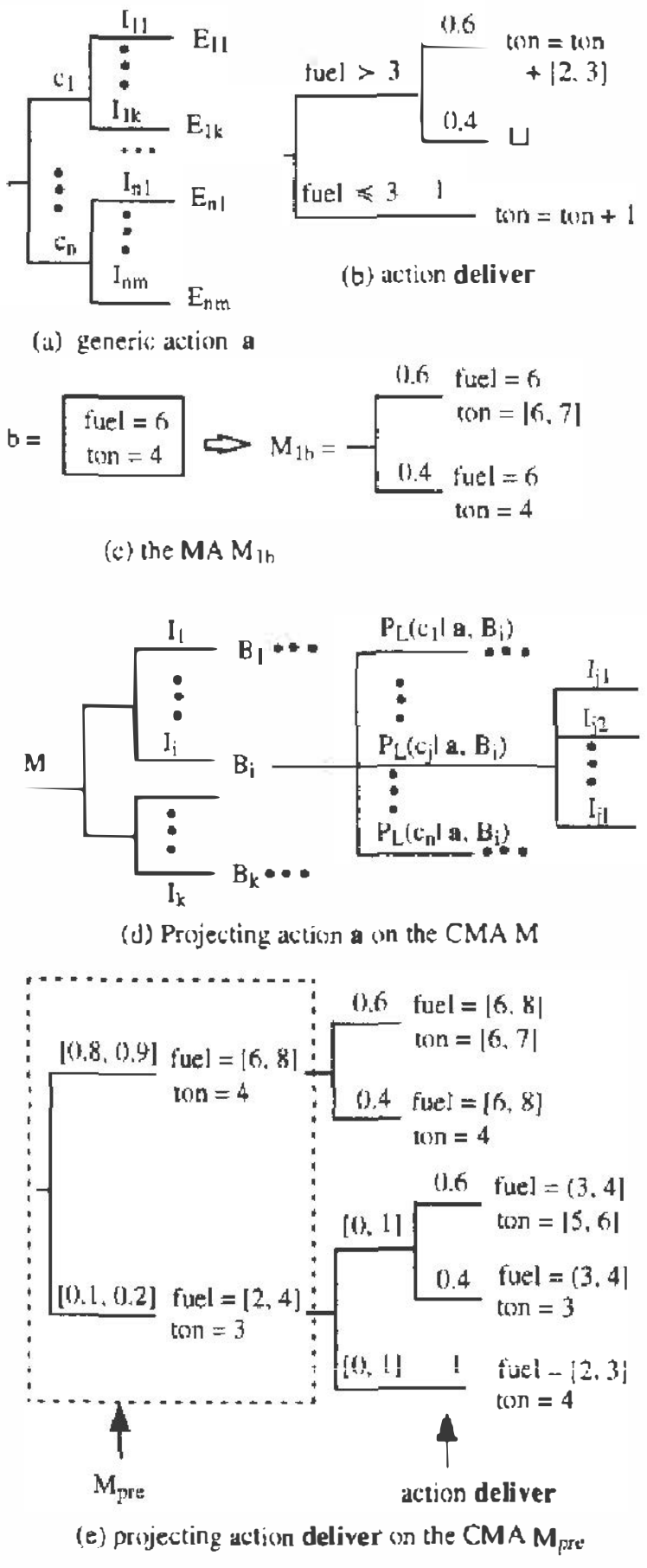

Figure 2: Examples of action and projecting action.

We can say that each condition $c_i$ is associated with an IMA $M_i$ whose branches are $\langle I_{ij}, E_{ij} \rangle$; but to be more precise, $M_i$ should be called a *functional* IMA because its leaf nodes, effects $E_{ij}$, are functions instead of sets. For a state $b$ and a condition $c_i$ we write $M_{ib}$ to denote the instantiated IMA computed from $M_i$ by replacing each leaf node $E_{ij}$ in $M_i$ with the set of states $E_{ij}(b)$. Figure 2.2 shows the computed IMA $M_{1b}$ for the state $b$ and the condition $c_1 = (fuel > 3)$ of the action **deliver** depicted in Figure 2.b.

In contrast to the SPD model, in which executing an action $\mathsf{a}$ on a state $b$ yields a single probability distribution, in our model executing an action $\mathsf{a}$ on a state $b \in c_i$ yields the set of probability distributions represented by the IMA $M_{ib}$. This means that the true post-execution probability distribution is in the set $\wp(M_{ib})$.

Based on the above semantics the set of all possible probability distributions over $\Omega$ that can result from executing action a on a probability distribution $P_{pre}$ can be computed as $exec(\mathsf{a}, P_{pre}) = \{P_{post} | \forall\ A \subseteq \Omega\ \ P_{post}(A) = \sum_b P_{pre}(b) \cdot P_b(A); P_b \in exec(\mathsf{a}, b)\}$ where $P_{post}(A)$ and $P_b(A)$ are the probabilities of the set $A$ in the probability distributions $P_{post}$ and $P_b$, respectively.

When the world is represented with a CMA $M_{pre}$, executing action a on this world yields the set $exec(\mathsf{a}, M_{pre}) = \bigcup_{P_{pre} \in M_{pre}} exec(\mathsf{a}, P_{pre})$.

In our framework, abstract actions are created by performing abstraction operations on concrete or lower-level abstract actions. Abstract actions have a syntax similar to that of concrete actions except for the fact that the conditions of an abstract action are not required to be mutually exclusive. Abstract actions will be discussed in more detail in Section 4.

### Projecting Concrete Actions and Plans

We have developed a sound projection procedure called *CMA-project* to project actions and plans. To introduce this procedure, we need first to define the *loose conditional probability* $P_L(c_i|\mathsf{a}, B)$. Given a set $B \subseteq \Omega$, an action a, and a condition $c_i$ of this action, $P_L(c_i|\mathsf{a}, B)$ is defined to be (a) 0 if $B$ is not consistent with $c_i$ (i.e., $B \cap c_i = \emptyset$); (b) 1 if $B$ is consistent with $c_i$ ($B \cap c_i \neq \emptyset$) and is not consistent with any other condition of a; and (c) the interval [0,1] if $B$ is consistent with at least two conditions of a, including the condition $c_i$.

Given the above definition, the projection procedure can be informally described as follows: Suppose we want to project the generic action a shown in Figure 2.a on the CMA M shown in the left part of Figure 2.d. First, we "grow" the tree (of the CMA) M by attaching the tree of action a to every leaf node of the tree M. The right part of Figure 2.d shows how the tree of a is attached to the leaf node $B_i$ of the tree M. Next, for each subtree that has as its root a leaf node - say, $B_i$ - of the tree M we perform the following assignment and replacement process: We assign to each branch leading out from node $B_i$ and was previously associated with the condition $c_j$ of action a the branch probability interval $P_L(c_j|\mathsf{a}, B_i)$; and we replace each node previously associated with an effect $E_{jq}$ of a with a node representing the set of states $E_{jq}(B_i \cap c_j)$. Finally, for any index pair $i$ and $j$ if we have $P_L(c_j|\mathsf{a}, B_i) = 0$ then we prune away the branch leading out from the node $b_i$ and is associated with condition $c_j$; we also prune away the subtree that is attached to this branch. The resulting tree "contains" the tree M and represents the CMA that is the result of projecting the action a on the CMA M.

A concrete example of projecting an action is shown in Figure 2.e. Here the action **deliver** is projected on the CMA $M_{pre}$; the post-projection CMA is the CMA $M_1$ shown in Figure 1.d.

Regarding the soundness of the above projection procedure we have the following theorem

**Theorem 1** *The projection procedure CMA-project is sound. That is, for any concrete plan p and world w we have* $CMA-project(p, w) \supseteq exec(p, w)$.

### Time Analysis of The CMA Framework

The CMA framework is a generalization of the SPD framework from the perspective of expressiveness. SPD worlds are IMAs with the $I_i$ being point intervals and the $B_i$ being singleton sets. SPD actions are special cases of CMA actions. When a CMA action a and CMA world $M$ can also be represented in the SPD framework, projecting a on $M$ using the procedure *CMA-project* is the same as using forward projections in the SPD framework (see [11] for descriptions of SPD forward projection procedures).

In the CMA framework projection complexity can be roughly measured with the number of nodes in the tree of the last CMA that results from projection. Assume the initial world is a singleton IMA - an IMA having a single branch, each action has $k$ branches associated with each condition, and each plan is a sequence of $n$ actions. Assume further that when projecting an action on a set of states, $t$ conditions are consistent with this set. Then the number of nodes in the final CMA that results from projection is $((tk)^{n+1} - 1)/((tk) - 1))$ Because of added increase in the expressive power, any complexity comparison between the SPD and CMA frameworks is skewed save for the case $t = 1$. In this case it is easy to see that the two projection complexity is the same. In general, the projection time in the CMA framework is a $t^n$ factor greater than that in the SPD framework.

## 4 ABSTRACTING ACTIONS

From the time analysis of plan projection in the previous section it is clear that the complexity of plan projection depends on two major factors: (1) the number of actions in a plan and (2) the number of branches in an action. When plan projection is used in the process of finding the best plans from a given set of plans, the complexity of planning also depends on the number of plans in the set. In a previous paper on action abstraction [8] we have indentified three types of sound abstraction that help reduce the complexity of planning: *intra-action, inter-action,* and *sequential* abstraction. Intra-action abstraction reduces the number of branches in an action by abstracting a set of branches in this action into an abstract branch. Inter-action abstraction abstract a set of alternative actions[3] into an abstract action with the same branching factor. This abstraction type is used to abstract a set of plans, thus reducing the number of plans. Sequential abstraction abstract a sequence of actions into an abstract action, thus reducing the number of actions in a plan.

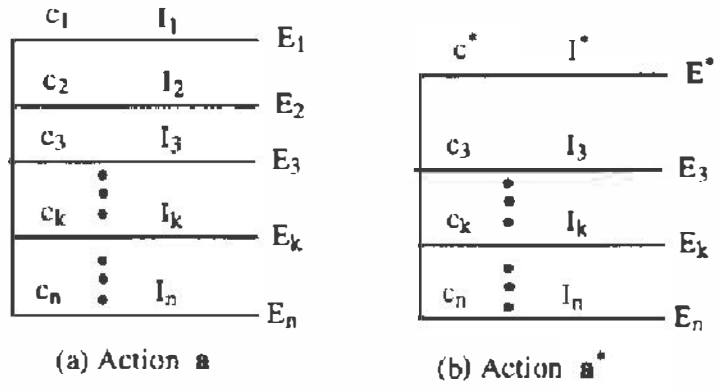

Figure 3: Intra-action abstraction.

In this section we present abstraction procedures for the three abstraction types mentioned above. Due to space limitation we present only the abstraction procedures, but see [3] for examples and a detailed discussion of the derivation process (which is technically quite complicated).

Since the description of an abstract action may not be describable in terms of conditions associated with IMAs, in the rest of this paper we will refer to a concrete or abstract action as a finite set of branches (triples) $\langle c_i, I_i, E_i \rangle$, where the conditions $c_i$ may not be mutually exclusive but are jointly exhaustive.

**Intra-Action Abstraction Procedure**

Intra-action abstraction is the process of creating an abstract action a* out of an action a by bundling a number of branches of a into an abstract branch of a* while transferring the remaining branches of a intact into a*.

Let the branches of a be $\langle c_1, I_1, E_1 \rangle, \cdots, \langle c_n, I_n, E_n \rangle$ (Figure 3.a). Without loss of generality assume that we are to abstract the two branches $\langle c_1, I_1, E_1 \rangle$ and $\langle c_2, I_2, E_2 \rangle$ into the abstract branch $\langle c^*, I^*, E^* \rangle$ of a* (Figure 3). We specify $c^*$ to be the disjunction $c_1 \vee c_2$, and create as the effect the function $E^* : \Omega \rightarrow 2^\Omega$ such that for all $B \subseteq \Omega$, $E^*(B) \supseteq E_1(B)$ and $E^*(B) \supseteq E_2(B)$. The interval $I^*$ is a number and is specified as $\max\{I_1, I_2\}$ in the case $c_1 \cap c_2 = \emptyset$, and as $\max\{I_1 + I_2\}$ otherwise. The remaining branches of a are transferred intact into a* (Figure 3.b).

**Inter-Action Abstraction Procedure**

Inter-action abstraction is the process of creating an abstract action a* out of a set of alternative actions $\{a_1, a_2, \cdots, a_n\}$.

To simplify the presentation of the abstraction procedure we are considering only the case of abstracting two actions $a_1$ and $a_2$ into an abstract action a*. Let the branches of $a_1$ be $\langle c_{1i}, I_{1i}, E_{1i} \rangle$, $i = 1, 2, \cdots, m$; and the branches of $a_2$ be $\langle c_{2j}, I_{2j}, E_{2j} \rangle$, $j = 1, 2, \cdots, n$. Assume that $m \geq n$. These two actions are depicted in Figures 4.a and 4.b. We abstract $a_1$ and $a_2$ by pairing each branch of $a_1$ with a branch of $a_2$. In case the numbers of branches differ ($m > n$), we add $m - n$ additional branches with condition "false" and probability interval zero to the action $a_2$ to make the numbers equal. In Figure 4.b these $m - n$ additional branches are depicted with dotted lines. Without loss of generality, assume that we pair and abstract the first branch of $a_1$ with the first branch of $a_2$ into the first branch of the abstract action, the second branch of $a_1$ with the second branch of $a_2$ into the second branch of the abstract action, and so on. The resulting abstract action a* is depicted in Figure 4.c.

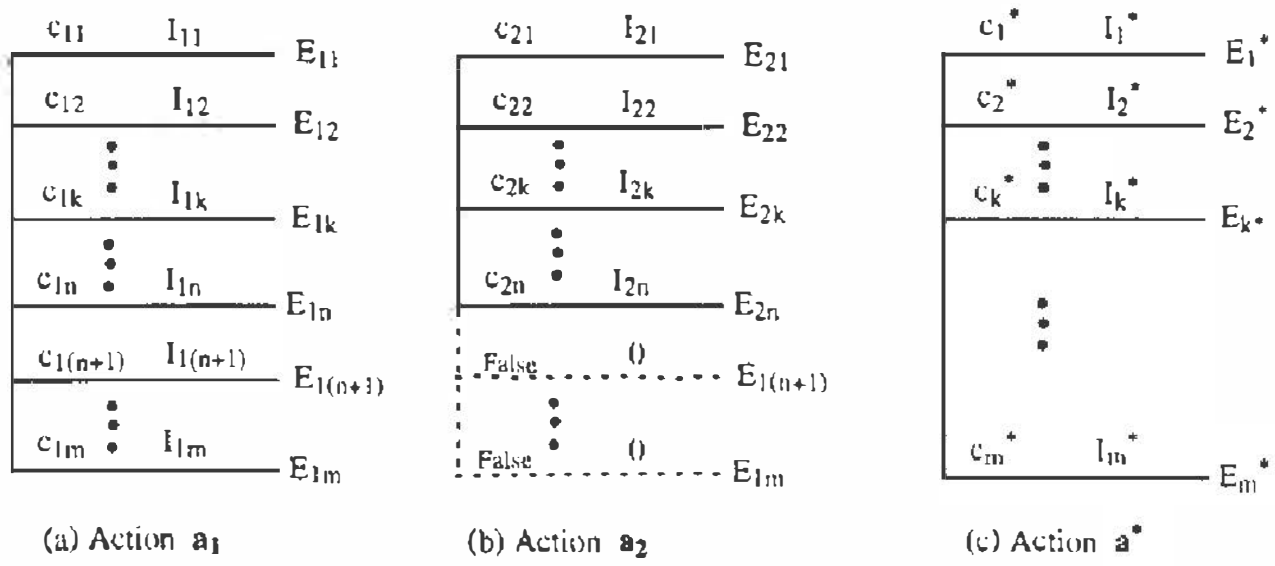

Figure 4: Inter-action abstraction.

For each set $S = \{\langle c_{1k}, I_{1k}, E_{1k} \rangle, \langle c_{2k}, I_{2k}, E_{2k} \rangle\}$ we create an abstract branch $\langle c_k^*, I_k^*, E_k^* \rangle$ of a* corresponding to $S$ by choosing $c_k^* = c_{1k} \vee c_{2k}$, $I_k^* = [\min\{I_{1k}, I_{2k}\}, \max\{I_{1k}, I_{2k}\}]$ and $E_k^*$ such that for all $B \subseteq \Omega$ $E_k^*(B) \supseteq E_{1k}(B)$ and $E_k^*(B) \supseteq E_{2k}(B)$.

**Sequential Abstraction Procedure**

Sequential abstraction is the process of creating an abstract action a* out of a sequence of action $a_1 a_2 \cdots a_n$.

Consider the case where a sequence of two actions $a_1 a_2$ is abstracted into a*. Let the branches of $a_1$ be $\langle c_{11}, I_{11}, E_{11} \rangle, \cdots, \langle c_{1m}, I_{1m}, E_{1m} \rangle$; and assume that action $a_2$ has $n$ branches, subscripted in a similar manner (Figures 5.a and 5.b). We shall abstract sequence $a_1 a_2$ by pairing every branch of $a_1$ with every branch of $a_2$ and creating an abstract branch of a* out of each of these sets.

Consider abstracting the pair: $\langle c_{11}, I_{11}, E_{11} \rangle$ of $a_1$ and $\langle c_{21}, I_{21}, E_{21} \rangle$ of $a_2$ into $\langle c_1^*, I_1^*, E^*{}_1 \rangle$ of a*. Let $c_2' = \{b \in B | E_1(b) \cap c_2 \neq \emptyset\}$. We choose $c^*$ to be the conjunction $c_1 \wedge c_2'$, $I^*$ to be the interval $[\min I_1 \cdot \min I_2, \max I_1 \cdot \max I_2]$, and $E^*$ to be a function $E^* : \Omega \rightarrow 2^\Omega$ such that $E^*(B) \supseteq E_2(E_1(B))$ for all $B \subseteq \Omega$.

In practice $c_2'$ is often a logical sentence derived from $c_2$ and the description of function $E_1$. If we can conclude that the sentence $c_1 \wedge c'_2$ is false then we do not need to create the branch $\langle c^*, I^*, E^* \rangle$ corresponding to the set with branches $c_1$ of $a_1$ and $c_2$ of $a_2$.

[3]Alternative actions are those that are in the same planning stage and so are interchangeable.

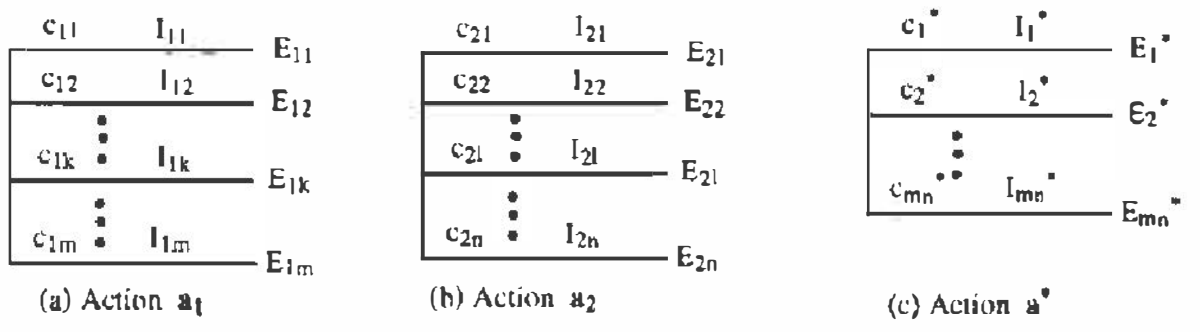

Figure 5: Sequential abstraction.

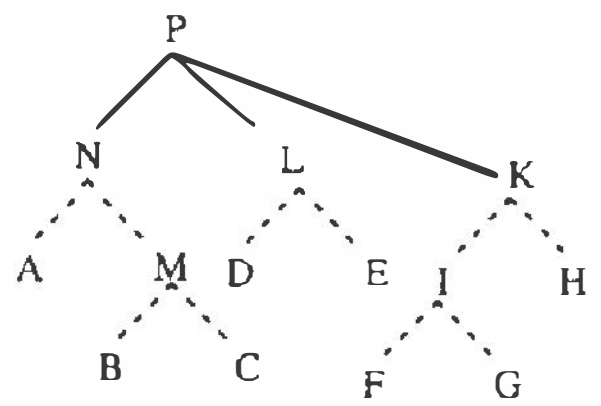

Figure 6: An abstraction hierarchy.

**Projecting Abstract Actions and Plans**

The procedure *CMA-project* presented in Section 3 can be used to project abstract actions and plans. This is possible because the syntax of abstract actions differs from that of concrete actions only in the fact that the conditions of an abstract action are not required to be mutually exclusive; and the projection procedure does not rely on this syntax difference.

**Soundness of Abstraction Procedures**

We can model the process of repeatedly using the three presented abstraction procedures on a set of actions with an *abstraction hierarchy*, which is a tree describing how an action represented by a parent node is abstracted from the actions represented by the direct child nodes. Figure 6 shows an abstraction hierarchy; here the dashed lines encode inter-action abstraction relationships – for example, the action M is inter-abstracted from the actions B and C, and the solid lines encode sequential abstraction relatinships – for example, the action P is sequentially abstracted from the actions N, L, and K.

Using an abstraction hierarchy we can find all concrete instantiations of an action. For example, in the abstraction hierarchy shown in Figure 6 the concrete instantiations of the action K are F, G, and H; the concrete instantiations of the top action P are the eighteen plans ADF, ADG, ADH, AEF, ..., AEH, BDF, etc. Similarly, for any abstract plan encoded in this abstraction hierarchy we can find all of its concrete instantiations. For example, a concrete instantiation of the plan NLK is ADF.

Regarding the soundness of our abstraction framework we have the following theorem

**Theorem 2** *Abstraction using the presented intra-, inter-, and sequential abstraction procedures are sound. That is, for any CMA M, plan p, and its concrete instantiation p', we have*
$CMA - project(p, M) \supseteq exec(p', M)$.

## 5 DISCUSSIONS AND RELATED WORKS

The work presented here is a significant extension of two previous papers on the same subject [8, 4]. In [8] we derived intra- and inter-abstraction methods for action abstraction in the SPD framework. We proved the soundness of the derived abstraction procedures. Unfortunately, the post-projection result can not be represented with a single probability distribution, so a sequence of actions cannot be projected.

Our attempt at developing a more expressive representation framework that can accommodate sound abstraction resulted in the IMA framework presented in [4]. This framework differs from the CMA framework in that we use IMAs to represent worlds. So even though we use essentially the same projection rule as *CMA-project* to project actions, after each action projection the resulting CMA tree must be "flattened" into an IMA. This "flattening" process poses two problems: First, we lose the probability constraints encoded in the tree structure. As a result, after projecting an action, the post-projection set is a rather *loose* convex set that subsumes the post-execution set of probability distributions. Second, the procedure that we use to compute the expected utility of a plan is not structured with respect to actions in the plan because the "flattening" process destroys whatever structure information we have. This loss of structuredness makes estimating information loss due to abstraction a nearly impossible task.

Our current work addresses these problems by adopting the CMA framework to represent worlds, thus avoiding "flattening" trees. As a result, the post-projection set of probability distributions in the CMA framework is a convex set that subsumes the post-execution set of probability distribution and *is subsumed* by the post-projection convex set obtained in the IMA framework through a similar projection process. In [5] we show that in the general case the post-projection convex set in the CMA framework is not the *smallest* convex set that subsumes the post-execution set. We haven't been able to show in general how loose the post-projection set is compared to the post-execution set, but see [5] for some partial results, as well as preliminary results on estimating utility loss due to abstraction and on methods to construct abstraction hierarchies.

We have implemented the DRIPS planner that searches through a plan space to find the plan with the highest expected utility [7]. DRIPS uses the abstraction techniques discussed in this paper to avoid explicitly examining every individual plan. We have applied DRIPS to a variety of problems in medical decision making with good results [7]. For example in a domain of 6,206 plans, DRIPS evaluates less than 600 plans to find the optimal plan. This result suggests that the abstraction techniques, even though preliminary, are applicable in solving real-world problems.

We have used a set-relational approach in defining CMAs as representing sets of probability distributions. For a different approach that uses *affin-operators* to construct CMAs, see [6]. In that paper the authors present essentially the same world representation, but a slightly different action representation, projection procedures, and action abstraction procedures. The presented abstraction procedures are weaker forms of the abstraction procedures presented in Section 4.

Our work on intra-action abstraction draws ideas from the work of Hanks [9]. In that work he suggested "bundling" action branches (i.e., intra-action abstraction) to reduce the complexity of projection. The BURIDAN planner [11] implemented this bundling technique in one of its projection algorithms. No systematic procedure for bundling branches was presented.

Recently, Smith *et. al.*[13] proposed reducing the structural complexity of a plan represented as a directed diagram by replacing a sequence of actions with an aggregate action the effect of which is consistent with the end effect of the action sequence. This amounts to doing sequential abstraction in our framework. Their work, however, did not offer any formal abstraction theory or concrete application.

Horvitz [10] also discussed utility-based state and action abstraction. In his framework abstraction is approximate, and utility is action-based, instead of plan-based.

Boutilier *et. al.* [1] proposed an approximate abstraction method for Markov Decision Processes. In his method the complexity of the domain is reduced by first identifying a set of "irrelevant" domain attributes; abstract states and actions are then created by ignoring these attributes. The new abstract problem has fewer states and actions, and therefore can be solved faster.

**Acknowledgments**
We thank Richard Goodwin and Vu Ha for valuable comments on parts of this paper. This work was partially supported by NSF grant IRI-9509165, a Sun Microsystems AEG award, and a Graduate Fellowship from the University of Wisconsin-Milwaukee.